# Robustness and Visual Explanation for Black Box Image, Video, and ECG Signal Classification with Reinforcement Learning

**Soumyendu Sarkar**[*][†]**, Ashwin Ramesh Babu**[†]**, Sajad Mousavi**[†]**, Vineet Gundecha, Avisek Naug, Sahand Ghorbanpour**

Hewlett Packard Enterprise

{soumyendu.sarkar, ashwin.ramesh-babu, sajad.mousavi, vineet.gundecha, avisek.naug, sahand.ghorbanpour}@hpe.com

## Abstract

We present a generic Reinforcement Learning (RL) framework optimized for crafting adversarial attacks on different model types spanning from ECG signal analysis (1D), image classification (2D), and video classification (3D). The framework focuses on identifying sensitive regions and inducing misclassifications with minimal distortions and various distortion types. The novel RL method outperforms state-of-the-art methods for all three applications, proving its efficiency. Our RL approach produces superior localization masks, enhancing interpretability for image classification and ECG analysis models. For applications such as ECG analysis, our platform highlights critical ECG segments for clinicians while ensuring resilience against prevalent distortions. This comprehensive tool aims to bolster both resilience with adversarial training and transparency across varied applications and data types.

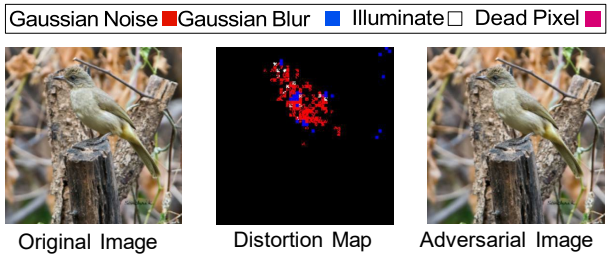

Figure 1: Mix of distortion filters for Adversarial Attack. Steps: 176, L2: 4.45. Demo: https://tinyurl.com/24ww544s

## Introduction

Deep learning models, despite their prowess, are vulnerable to input data corruption, posing challenges in safety-critical applications like self-driving cars and facial recognition. Black-box attacks generally work with limited model information but tend to be inefficient, relying heavily on hand-crafted heuristics (2; 1; 24). There has been substantial progress in Reinforcement Learning across various fields, from complex control systems to training LLMs (19; 3; 6; 14; 12; 15; 16; 13; 18; 17; 23; 9; 21; 4). These smart agents can navigate through various environments and take a sequence of actions to converge on the goal with a long-term view, which makes them very powerful. Addressing these issues, we introduce a Reinforcement Learning agent for a Platform (RLAB) capable of efficient adversarial attacks. This agent employs a "Bring Your Own Filter" (BYOF) approach (figure 1) and utilizes a dual-action mechanism to manipulate image distortions, with the aim of high success rates with fewer queries. For each application, we evaluated the performance of the proposed frameworks with various models and data sets to show the reliability of our method. We consider three types of metrics to evaluate performance, and our results show that the proposed reinforcement learning-based attack strategy generates superior results in all three applications compared to the state-of-the-art approaches. The main contributions of this work are:

- A **common attack framework that spans multiple dimensions, from 1D ECG signals to 2D images, and 2+D videos**
- Reinforcement Learning-based Adversarial Attack with multiple custom distortion types to measure the lowest distortion needed for misclassification as a metric for robustness and resiliency.
- Visual explanation in the form of localization and heat map derived from RL attack agent.
- Adversarial training to enhance robustness.

## Proposed Method

### Problem Formulation

A trained Deep Neural Network (DNN) model under evaluation can be represented as $y = argmax f(x; \theta)$. Our approach generates perturbation $\delta$ such that, $y \neq f(x + \delta; \theta)$. The distance between the original and the adversarial sample, $D(x, x + \delta)$ will be any function of the $l_p$ norms. The objective is to fool the classifier while keeping $D$ to a minimum.

### Robustness Evaluation

The input data are divided into fixed size patches of size $n$ for 1D, $n \times n$ for 2d data, $t \times n \times n$ for 3D data where t represents the temporal dimension. For every step, the RL agent decides to take two actions,

1. Patches to which distortions are added

---

[*]Corresponding Author
[†]These authors contributed equally.

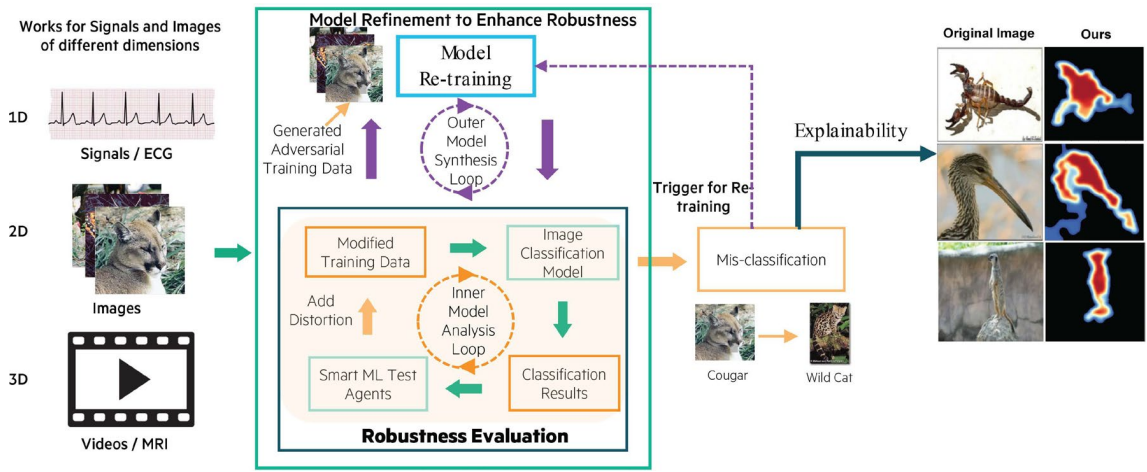

Figure 2: Overview of RL Framework for Robustness and Explainability for signals, images, and video.

2. patches from which distortions are removed

This process is done iteratively until the model misclassifies the data or until the budget for the number of maximum allowed steps is reached. This loop is represented as the **"robustness evaluation"** block in the figure 2. The intuition behind having two actions (addition and removal) is inspired by the application of reinforcement learning for board games where the most effective moves or actions can be determined using methods such as Deep Tree Search (DTS) (22). Unlike board games, there is a possibility to reset earlier actions that were taken in the past that proved to be less effective, reducing the computational complexity from $O(N^d)$ to $O(N)$. Here, $N$ represents the computational complexity of one level of evaluation and corresponds to the size of the data, and $d$ represents the depth of the tree search which translates to the number of actions taken ahead of time. The generated adversarial samples are further used to fine-tune the victim model to improve robustness which is represented in the figure 2 as **"model refinement to enhance robustness"**.

**Bring Your Own Filter**

The RLAB platform is extremely versatile with any type of distortion of choice. The RL algorithm learns a policy to adapt to the filter used such that the adversarial samples are generated with minimum distortion $D$. Furthermore, the algorithm can be used with a mixture of filters such that the agent first decides which filter to use for every step and further determines the patches to which the distortion should be added. We experimented with four naturally occurring distortions (Gaussian Noise, Gaussian blur, dead pixel, and illuminate).

**Explainability**

The RL agent has been trained to add distortion to the most sensitive region of the data such that the misclassification can be introduced with a minimum number of steps. This approach has encouraged the agent to add distortion to the region of the data that corresponds to the predicted class. This creates an accurate localization of the objects/peaks in the scene/signal, which is represented in figure 2 as "explainability".

**Results and Discussion**

For all three applications (ECG analysis, Image Classification, Video Classification), we use three evaluation metrics, average success rate, number of queries, and the $l_2$, $l_{inf}$, to measure the effectiveness of the attack framework. For all applications, we have evaluated more than 1 dataset and more than three different victim models to assess the effectiveness of the proposed framework. Results prove that the RL agent could generate an "average success rate" of 100 percent most of the time with a much smaller query budget when compared to the competitors (8; 10; 11; 16). Furthermore, the proposed framework could maintain the "average number of queries" lower than the competitors for all three applications. Also, the effectiveness of localization is evaluated with metrics such as dice coefficient and IOU and compared with the popular gradient and non-gradient-based approaches (20; 5; 7), with the proposed method showing superiority over the other approaches. Also, retraining the model with adversarial samples significantly improved robustness when evaluated on benchmark datasets (10).

**Conclusion**

The proposed reinforcement learning-based attack framework is effective causing misclassification for many applications with different data dimensions, showing its ability to generalize for different data dimensions. The approach is capable of using any distortion types that suit the use case to generate meaningful adversarial samples. Furthermore, the visual explanations generated by the RL agents provide insights into the decisions of the AI models. The framework is currently being evaluated for LLMs.

**References**

[1] Maksym Andriushchenko, Francesco Croce, Nicolas Flammarion, and Matthias Hein. Square attack: a